\documentclass[letterpaper, 10 pt, conference]{ieeeconf}  

\IEEEoverridecommandlockouts                              

\usepackage{graphics} 
\usepackage{epsfig} 
\usepackage{times} 
\usepackage{amsmath} 
\usepackage{amssymb}  
\usepackage{hyperref} 
\usepackage{caption}
\usepackage{subcaption}
\usepackage{booktabs}
\usepackage{multirow}
\usepackage{multicol}
\usepackage{color}
\let\labelindent\relax
\usepackage{enumitem}
\usepackage{float}
\usepackage{boldline}
\usepackage{cite}
\usepackage{tablefootnote}
\usepackage{hyperref}
\usepackage{url}

\title{\LARGE \bf
MUSE: Multimodal Uncertainty Quantification of State Estimation
}

\author{Minkyung Kim$^{*1}$, Henry Che$^{*2}$, Bhargav Chandaka$^2$, Bhumsitt Pramuanpornsatid$^2$, Chengyu Yang$^1$, \\ Sheng Cheng$^1$, Xiaofeng Wang$^3$, Naira Hovakimyan$^1$, Shenlong Wang$^2$
\thanks{$^*$These authors contributed equally to this work.} 
\thanks{$^1$Authors are with the Department of Mechanical Science and Engineering, University of Illinois Urbana-Champaign, Champaign, IL 61801, USA. ({\tt\small email: \{mk58,cy45,chengs, nhovakim\}@illinois.edu})}
\thanks{$^2$Authors are with Siebel School of Computing and Data Science, University of Illinois Urbana-Champaign, Champaign, IL 61801, USA. ({\tt\small email: \{hungdc2, bhargav9, bp17, shenlong\}@illinois.edu})}
\thanks{$^3$Author is with the Department of Electrical Engineering, University of South Carolina, Columbia, SC 29208, USA. ({\tt\small email: wangxi@cec.sc.edu})}
}

\begin{document}

\maketitle
\thispagestyle{empty}
\pagestyle{empty}

\begin{abstract}
Accurate visual state estimation has been a central topic in robotics with a wide range of applications in robot navigation, autonomous driving, and autonomous flight. Recent advances in robot perception have led to significant improvements in the accuracy and robustness of state estimation, yet a fundamental challenge remains in how to quantify and calibrate its precision, i.e., how confident we are in an estimate and whether failures can be detected. This issue is particularly pronounced in visual–inertial odometry (VIO), where the heteroscedastic and multimodal nature of the problem makes uncertainty quantification especially difficult. This paper introduces MUSE (Multimodal Uncertainty Quantification of State Estimation), a novel real-time learning-based framework that leverages the strong and efficient sequential modeling capacity of Mamba to estimate localization uncertainty from multiple asynchronous sensor streams. Experiments on both public and in-house datasets demonstrate that MUSE achieves superior reliability and robustness compared to existing uncertainty quantification methods, and ablation studies justify the benefits of its key design choices. We release our source code and dataset at \url{https://github.com/hungdche/MUSE}.
\end{abstract}
\section{INTRODUCTION}

State estimation is fundamental in robotics, enabling applications ranging from autonomous driving and drone navigation to robot operation in GPS-denied environments. Research over the past few decades in perception-based state estimation, such as visual odometry (VO) and visual–inertial odometry (VIO), has yielded significant advances in estimating a robot's pose (position and orientation), making these methods indispensable for safety-critical tasks. However, accuracy alone is not sufficient: robotic systems must also reason about the reliability of their state estimates. In practice, odometry estimates remain prone to errors from sensor noise, environmental ambiguity, and algorithmic limitations, making uncertainty quantification essential for safe and reliable decision-making. 

\begin{figure}[t]
    \centering
    \includegraphics[width = \columnwidth]{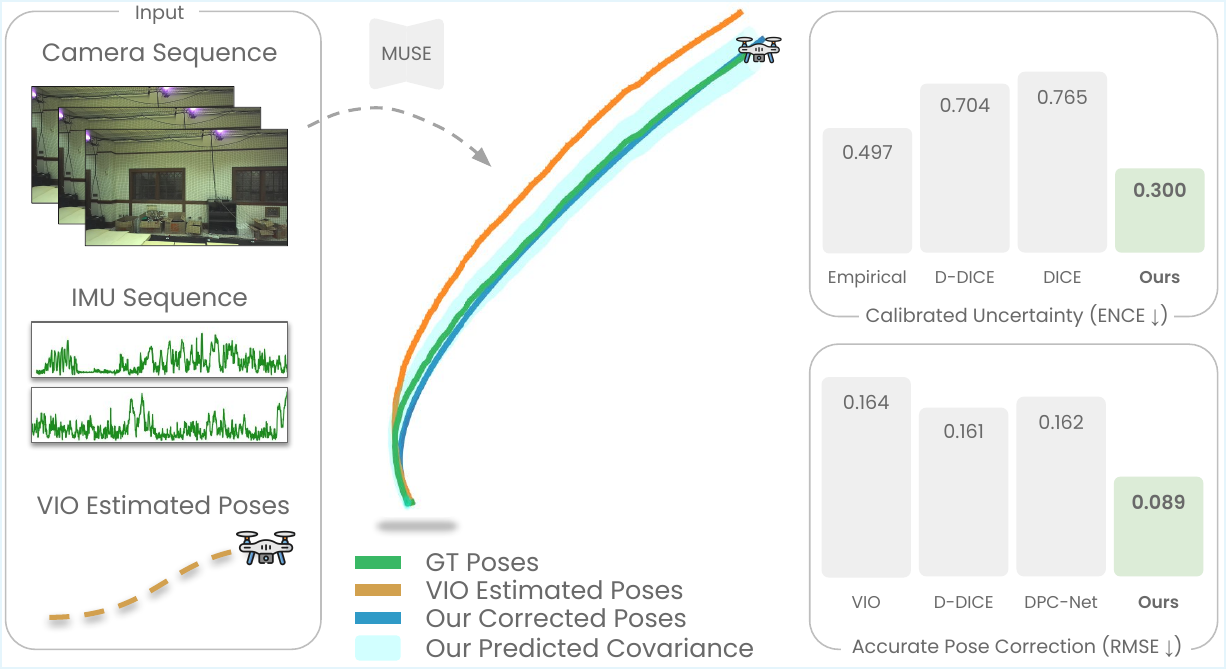}
    \caption{We present MUSE, a framework to jointly correct poses and predict heteroscedastic uncertainty across time. Our method is able to achieve improvement over various baselines in both pose correction accuracy and uncertainty.}
    \label{fig:teaser}
    \vspace{-0.7cm}
\end{figure}



VO and VIO approaches can be broadly categorized into three paradigms. Filter-based approaches~\cite{mourikis2007multi, sun2018robust} propagate covariance recursively but suffer from inconsistency due to linearization errors, model mismatch, drift, and limited belief representation. Optimization-based methods~\cite{engel2017direct, qin2017vins, qin2019a, mur2015orb, murORB2, campos2021orb} improve accuracy and robustness, yet estimate uncertainty post-hoc from the final solution, yielding local and poorly calibrated measures. Learning-based approaches~\cite{wang2017deepvo, wang2021tartanvo, teed2023deep} achieve state-of-the-art (SOTA) accuracy but remain overconfident, providing unreliable uncertainty estimates.

\begin{figure}
    \centering
    \includegraphics[width=\columnwidth]{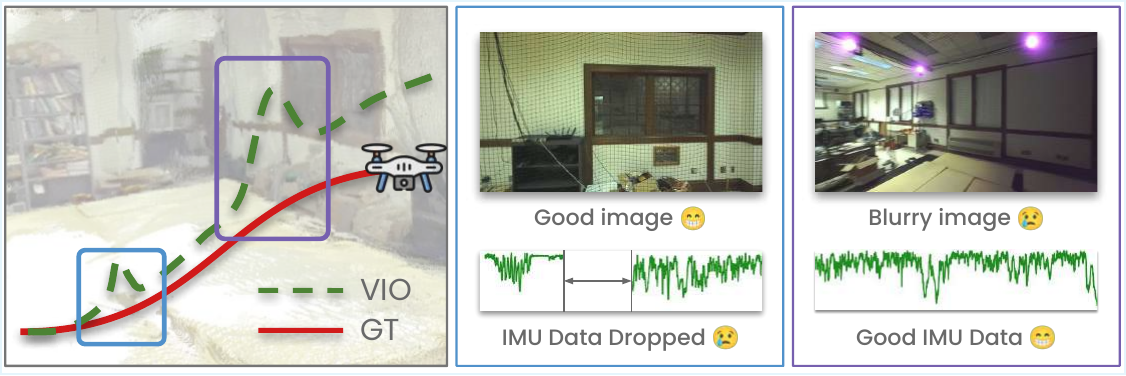}
    \caption{Motivation. Uncertainty and failures in visual–inertial odometry (VIO) arise from diverse and multimodal factors (e.g., IMU anomalies, sensor contamination). Our model is designed to learn to capture these multimodal cues for better uncertainty calibration. }
    \label{fig:multimodal}
\vspace{-6mm}
\end{figure}

Recent works attempt to learn pose uncertainty directly from sensory cues~\cite{liu2018deep, de2020simultaneously, peretroukhin2017dpc}. While effective to some extent, these methods have key limitations: they typically rely on instantaneous inputs without exploiting temporal information and can use only visual information as their primary input. Intuitively, as depicted in Fig. \ref{fig:multimodal}, pose drifts and uncertainty in VIO are usually more predictable via temporal and multimodal cues (e.g., sudden drop in IMU data, motion blur detected in images, or sudden sharp changes in raw odometry that disagree with IMU) rather than a single image cue. As a result, these methods' uncertainty predictions remain limited in both reliability and generality.


In this paper, we present {\bf MUSE}, a novel, real-time, and versatile framework for quantifying VO/VIO uncertainty from temporal multi-sensor streams. MUSE takes raw odometry and multi-sensor streams as input and predicts a \textit{non-zero-mean} Gaussian distribution over pose errors in SE(3), enabling simultaneous pose correction and calibrated uncertainty estimation. At its core, MUSE leverages Mamba \cite{gu2023mamba}, a powerful and efficient structured state-space model (SSM), to capture uncertainty over time windows and account for the heteroscedasticity of pose errors. \textit{To the best of our knowledge, this is the first framework that jointly addresses the multimodal, temporal, and heteroscedastic nature of odometry uncertainty.} Moreover, MUSE is directly deployable as a real-time plugin for any VO/VIO system.  


We validate MUSE on EuRoC~\cite{burri2016euroc} and our challenging in-house UnCal-Flight dataset across diverse VO/VIO frameworks. Results show improved pose correction and well-calibrated uncertainty estimates, outperforming various baselines (Fig.~\ref{fig:teaser}). A comprehensive ablation study further demonstrates the benefits of multimodal integration for robust uncertainty prediction.

Our main contributions are summarized as follows:  
\begin{enumerate}[label=(\roman*)]
    \item A novel architecture that learns multimodal, heteroscedastic uncertainty in odometry estimation.  
    \item An open-source, challenging drone navigation UnCal-Flight dataset for evaluating VO/VIO robustness.
    \item Extensive experiments on EuRoC and our dataset across multiple VIO modules, showing MUSE's effectiveness as a real-time plugin for existing systems.  
\end{enumerate}

\section{RELATED WORK}
\label{sec:Related}

\subsection{Visual(-Inertial) Odometry} 




Traditional VO/VIO pipelines include (i) filter-based estimators \cite{mourikis2007multi, sun2018robust} that fuse inertial and visual constraints online with recursive updates, and (ii) optimization-based estimators, such as VINS-Mono \cite{qin2017vins}, VINS-Fusion \cite{qin2019a}, and ORB-SLAM \cite{mur2015orb, murORB2, campos2021orb}, that refine a sliding window or global pose graph with geometric or photometric objectives. Kalman filter offers real-time efficiency but can suffer from inconsistency stemming from linearization, leading to overconfident or conservative covariances and drift. Optimization-based methods generally achieve higher accuracy, yet their uncertainty is typically local and not calibrated for predictive use, making it challenging to quantify heteroscedastic or multimodal error under changing conditions. 

Recently, deep-learning-based approaches (e.g., DeepVO \cite{wang2017deepvo}, TartanVO \cite{wang2021tartanvo}, DPVO \cite{teed2023deep}) achieve strong accuracy by directly estimating poses from visual inputs, but they frequently exhibit limited reliability in their uncertainty, especially under distribution shift or degraded sensing. To address this issue, unsupervised approaches like GANVO \cite{almalioglu2019ganvo} and SelfVIO \cite{almalioglu2022selfvio} remove the need for labeled trajectories by leveraging view-synthesis and adversarial training techniques. Meanwhile, Adaptive VIO \cite{pan2024adaptive} combines continual learning with traditional optimization to enable the system to adapt to the new environment. 

Despite these advancements, challenges remain in ensuring the reliability of VO/VIO under out-of-distribution or adverse conditions, including dynamic environments, sensor degradation, and low-light scenarios. Our work seeks to close these gaps by incorporating uncertainty quantification for a VO/VIO framework, which can be utilized alongside existing VO/VIO systems in safety-critical applications. 

\subsection{Uncertainty Quantification in SLAM}



Uncertainty quantification has been widely studied in machine learning~\cite{guo2017calibration} and is increasingly being addressed in VO/VIO through learning-based approaches. DICE \cite{liu2018deep} and D-DICE \cite{de2020simultaneously} utilize convolutional neural networks (CNNs) to learn a pose error model from raw images in the form of Gaussian distributions. DPC-Net \cite{peretroukhin2017dpc} predicts SE(3) pose error directly from images and performs corrections by employing pose graph relaxation techniques. More recent works \cite{dixit2024perceive, stutts2023lightweight} have explored statistically principled calibration, including conformal prediction frameworks that wrap pre-trained VO/SLAM to guarantee coverage. There are also research directions that seek to quantify heteroscedastic uncertainties when estimating poses. For instance, Uncertainty-Aware VO (UA-VO) \cite{costante2020uncertainty} provides confidence measures by accounting for epistemic and aleatoric uncertainties in VO. Similarly, D3VO \cite{yang2020d3vo}, a self-supervised method, incorporates predicted pose, depth, and photometric uncertainties. However, these visual-focused approaches typically do not consider uncertainties arising from other sensor modalities.

\begin{figure}[t]
    \centering
    \includegraphics[width = \columnwidth]{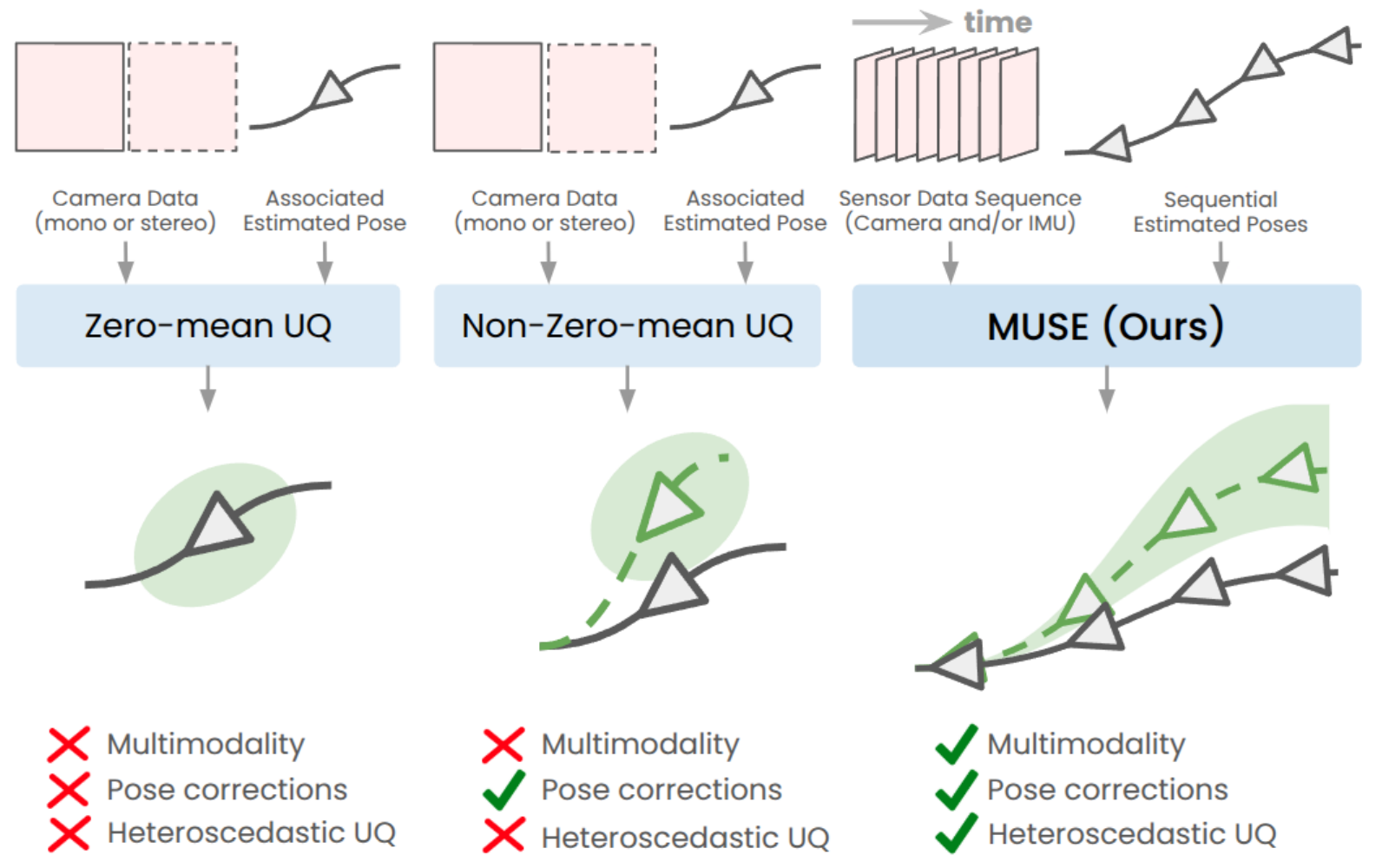}
      \caption{Comparison between prior uncertainty quantification (UQ) methods and MUSE. Our approach leverages multimodal sensor streams to capture the heteroscedastic nature of pose uncertainty, which prior methods cannot.}
    \label{fig:overview}
    \vspace{-0.5cm}
\end{figure}

Our work extends this line by explicitly targeting multimodal, sequential uncertainty quantification. Unlike prior visual-only-based or per-frame uncertainty predictors, we introduce a framework that fuses asynchronous visual, inertial, and odometry streams over long horizons. Our approach enables simultaneous pose correction and calibrated predictive uncertainty. This positions our method as the first to unify multimodal inputs, long-sequence modeling, and rigorous covariance prediction into a deployable plugin for existing VO/VIO systems, improving both accuracy and reliability in dynamic, resource-constrained environments.

\subsection{State-Space Models for Robotics}



While standard sequence models, such as Transformers \cite{vaswani2017attention}, excel at capturing complex relationships through self-attention, they suffer from quadratic computational complexity as sequence length increases. State Space Models (SSMs) have recently emerged as a powerful architecture inspired by classical dynamic systems. Early variants of SSM \cite{gu2021efficiently} first utilize specialized HiPPO parameterizations \cite{gu2020hippo}, enabling effective management of long-context information. Recently, Mamba \cite{gu2023mamba} introduced a selective mechanism, enabling the model to selectively emphasize or forget information, thereby significantly improving its expressive power.

These properties make SSMs particularly appealing for robotics, which relies on high-throughput sensory streams under real-time constraints. Applications in perception include multimodal temporal fusion for 3D perception \cite{luo2025overlapmamba} and adaptive parameter tuning in SLAM \cite{ma2025mamba}. Research in policy learning \cite{jia2024mail} shows that SSMs can replace heavier sequence models while preserving latency and robustness across long horizons. Frameworks like RoboMamba \cite{liu2024robomamba} further integrate Mamba into vision-language-action models, allowing robots to follow natural-language instructions grounded in visual context. Together, these advances highlight SSMs as a scalable backbone for real-time robotic systems that demand long-horizon and multimodal reasoning.

In this paper, we adopt Mamba \cite{gu2023mamba} as the backbone for quantifying uncertainty in localization. Its ability to selectively process information from long, complex sequences is ideal for efficiently quantifying uncertainty from demanding multi-sensor streams.

\section{Method}
\label{sec:method}

\begin{figure*}[!t]
    \centering
    \includegraphics[width = 0.9\textwidth]{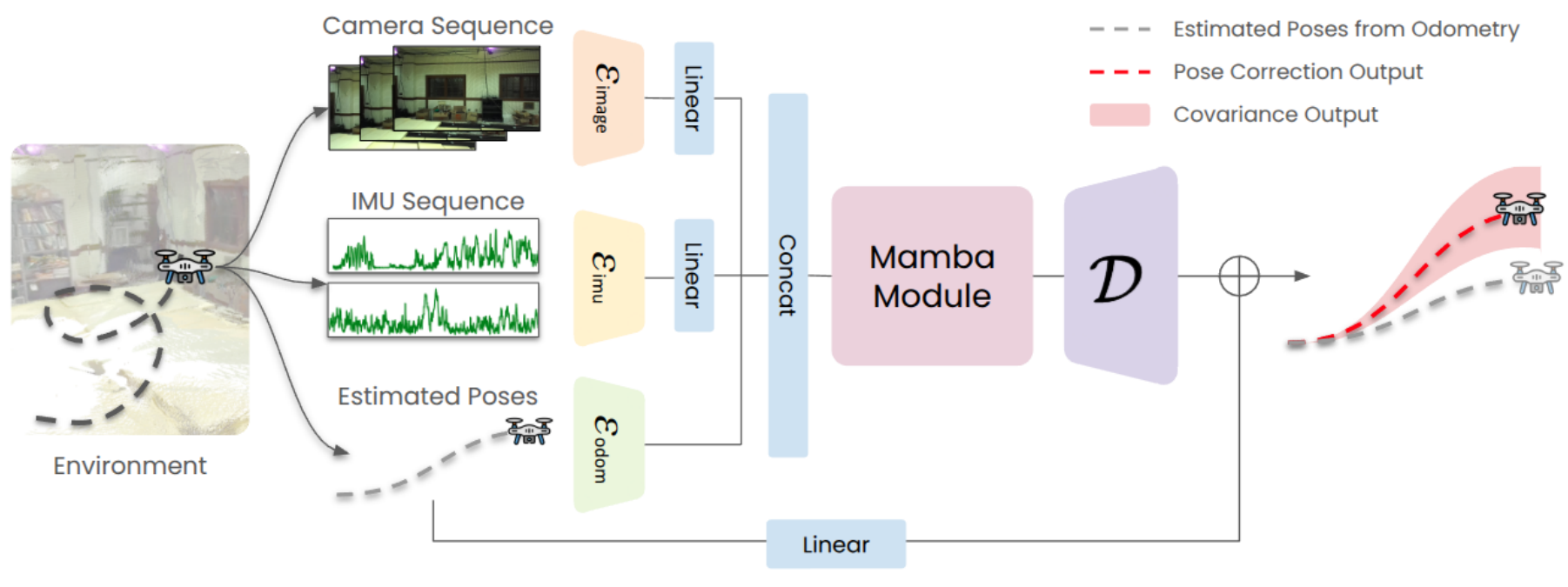}
    \caption{MUSE Architecture. MUSE takes multiple asynchronous sensor streams and the estimated pose of a given VO/VIO as the input. Each input stream (camera, IMU, and Odometry) is first encoded by its corresponding encoder before being concatenated together on the sequence length dimension. The output of concatenation is then passed through a Mamba module to capture the sequential relationship before being decoded into pose corrections and covariances.}  
    \label{fig: architecture}
    \vspace{-0.5cm}
\end{figure*}

\subsection{Preliminaries}
Let the ground‑truth pose at time~$i$ be $\mathbf T_{\mathrm{gt},i}\!\in\!\mathrm{SE}(3)$ and the estimated pose by an arbitrary VO/VIO module be $\hat{\mathbf T}_i$. Assume both trajectories share the same initial pose $\mathbf T_{\mathrm{gt},0}=\hat{\mathbf T}_0$. We define the rigid‑body pose estimator error at time~$i$ to be
\begin{equation}
    \mathbf T_{\mathrm{err},i}=\mathbf T_{\mathrm{gt},i}\,\hat{\mathbf T}_i^{-1}.
    \label{eq:T_err}
\end{equation}

Learning a parametric function to predict $\mathbf T_{\mathrm{err},i}$ is challenging and unstable due to the orthonormal constraints. Thus, we map the error $\mathbf T_{\mathrm{err},i}$ on SE(3) to the Lie algebra $\mathfrak{se}(3)$ via the logarithm map to obtain a six-value vector $\boldsymbol\xi_i=\log(\mathbf T_{\mathrm{err},i})\in\mathbb R^6$.  Throughout this paper, we adopt the ``hat'' operator $(\cdot)^{\wedge}\!:\!\mathbb R^6\!\to\!\mathfrak{se}(3)$ so that $\exp(\boldsymbol\xi_i^{\wedge})=\mathbf T_{\mathrm{err}, i}$. Following the prior work \cite{de2020simultaneously}, we model the error distribution as a non-zero-mean Gaussian distribution
\begin{equation}
    \boldsymbol\xi_i\sim\mathcal N(\boldsymbol\mu_i,\,\mathbf\Sigma_i)
\label{eq:distribution}
\end{equation}
with mean $\boldsymbol\mu_i\!\in\!\mathbb R^6$ and full covariance $\mathbf\Sigma_i\!\in\!\mathbb R^{6\times6}$. 

As $\mathcal N(\boldsymbol\mu_i,\,\mathbf\Sigma_i)$ is assumed to model the true underlying pose error distribution, we can use $\boldsymbol\mu_i$ to "correct" estimated error. We define the corrected pose $\hat{\mathbf T}'_i$ as follows: 
\begin{equation}
    \hat{\mathbf T}'_i=\exp(\boldsymbol\mu_i^{\wedge})\,\hat{\mathbf T}_i.
\label{eq:correct_pose}
\end{equation} Then, the error of the corrected pose $\mathbf T'_{\mathrm{err},i}$ is given by
\begin{equation}
    \mathbf T'_{\mathrm{err},i} 
    = \mathbf T_{\mathrm{gt},i}\,\hat{\mathbf T}'^{-1}_i
    = \mathbf T_{\mathrm{err},i}\,\exp(-\boldsymbol\mu_i^\wedge).
\end{equation} Let $\boldsymbol\xi'_i=\log(\mathbf T'_{\mathrm{err},i}) \in \mathbb R^6$.
Under the first-order Baker–Campbell–Hausdorff approximation, the error distribution of the corrected poses is approximately zero-mean with unchanged covariance:
\begin{equation}
\boldsymbol\xi'_i \sim \mathcal N(\mathbf{0}, \mathbf\Sigma_i) .
\label{eq:correct_cov_fixed}
\end{equation}


\subsection{Problem Definition}
Given a stream of multimodal sensor observations $\mathcal Z_t=\{ z_0,\dots, z_t\}$ where $z_i = \{z_\text{image}, z_\text{imu}\}_i$ and the sequence of estimated poses $\hat{\mathbf T}_{0:T}$ from a given odometry module over $T$ period, we seek to learn the distribution of the error between $\hat{\mathbf T}_{0:T}$ and the ground-truth poses $\mathbf T_{gt,0:T}$ through a parametric function $f_{\theta}$, such that:
\begin{equation}
    \boldsymbol\mu_{0:T}, \boldsymbol\Sigma_{0:T} = f_\theta(\mathcal Z_T, \hat{\mathbf T}_{0:T}),
\end{equation}
We expect our predicted 
$\boldsymbol\mu_{0:T}$ and $\boldsymbol\Sigma_{0:T}$ to accurately capture the true error distribution, where the mean provides pose correction and covariance offers posterior confidence. In this sense, our output is greatly desired for uncertainty-aware downstream planner or utilized as introspective model to improve perception accuracy. 

\subsection{Input Feature Extractions}
In light of our motivation to utilize multimodal data (Fig. \ref{fig:multimodal}), our pipeline takes as input multi-sensor data and raw odometry, which we process as follows.

\subsubsection{Sensor Inputs}
We consider stereo images and IMU data (linear accelerations and angular velocities) as the primary observational inputs of our model. We further remark that, owing to its modular design, MUSE can also support different sensor modalities and configurations. For visual inputs, we extract features using a pretrained \textit{SuperPoint} \cite{detone2018superpoint} encoder $\mathcal{E}_\text{image}$ (operating on grayscale images converted from RGB) and project them to $d_\text{image}$ via a small MLP. For inertial measurement data, we sample them at 200 Hz, truncate them into chunks based on timestamps of every successive estimated poses, and encode them with a pretrained \textit{RONIN} \cite{herath2020ronin} encoder $\mathcal{E}_\text{imu}$. Similarly, an MLP is employed to project the output features to  $d_\text{imu}$. 

\subsubsection{Odometry Input}
As our model is designed to correct the VIO estimates, learning the high-dimensional underlying distribution of the estimates is valuable. The upstream VO/VIO engine supplies each pose and, when available, the associated \(6\times 6\) covariance (e.g.\ MSCKF \cite{sun2018robust}). We encode these quantities with a single‑layer fully‑connected network (FCN) $\mathcal{E}_\text{odom}$, where the resulting feature dimension is $d_\text{odom}$.



After extracting the features from each input modality, we concatenate them along the sequence length dimension. Thus, the resulting feature is obtained via:
\begin{equation}
    f_\mathrm{input} = \mathcal{E}_\text{image}(\mathcal{Z}_\text{image})\oplus \mathcal{E}_\text{imu}(\mathcal{Z}_\text{imu}) \oplus \mathcal{E}_\text{odom}(\hat{T}_{0:T}),
\label{eq:features}
\end{equation}
where $\oplus$ denotes concatenation on the sequence dimension.
During training, the sensor encoders $\mathcal{E}_\text{image}$ and $ \mathcal{E}_\text{imu}$ are frozen for efficiency, while the projection layers and $\mathcal{E}_\text{odom}$ are updated by backpropagation. 



\subsection{Learning Error Distribution} 
\label{subsec:mamba}
To efficiently learn the sequential nature of pose estimation, we employ the recently published state-space model Mamba \cite{gu2023mamba} at the heart of our pipeline. The features obtained from Eq. \eqref{eq:features} are passed through a stack of Mamba blocks to learn the temporal correlations of the input modalities. For our Mamba module, we use $N$ blocks of Mamba, each consisting of a small linear layer followed by a Conv1D, SiLU activation, SSM \cite{gu2023mamba}, and a small linear layer. A residual connection, which consists of another linear layer and a SiLU activation, is then multiplied element-wise with the output of SSM, following \cite{gatedmlp}. In all experiments, we fix $N=4$ as a balance between efficiency and performance.

\subsection{Predicting Pose Corrections and Uncertainty} 
The desired output of our model is the mean $\boldsymbol\mu_i \in \mathbb{R}^6$ and covariance $\boldsymbol\Sigma_i\in\mathbb{R}^{6\times6}$ for $i \in [0,1,\cdots, T]$. However, directly predicting $36$ parameters representing $\boldsymbol\Sigma_i$ would lead to training instability due to the positive semi-definite constraints. Thus, following \cite{liu2018deep, de2020simultaneously}, we opt to reconstruct $\boldsymbol\Sigma_i$ through LDL Decomposition $\boldsymbol\Sigma_i = L_iD_iL_i^T$ by predicting 6 parameters $d_i$ and 15 parameters $l_i$ such that 
\begin{equation}
    \boldsymbol\Sigma_i = L_iD_iL_i^T = \mathrm{trilu}(l_i)\mathrm{diag}(\mathrm{exp}(d_i))\mathrm{trilu}(l_i)^T.
\label{eq:ldl}
\end{equation}Here, $\mathrm{diag}(\mathrm{exp}(d_i))$ denotes the $6 \times 6$ diagonal matrix, where $\mathrm{exp}(d_i) \in \mathbb{R}^6$ contains its diagonal entries, and $\mathrm{trilu}(l_i)$ denotes the unit lower triangular $6\times 6$ matrix with $l_i\in\mathbb{R}^{15}$ being the vector of nonzero and off-diagonal parameters. We apply the exponential function $\mathrm{exp}(\cdot)$ on $d_i$ to enforce positiveness and guarantee the existence and uniqueness of the decomposition. 

Implementation-wise, we employ two MLP decoders to predict (i) six elements of \(\boldsymbol{\mu}_i\) and (ii) 21 elements of $[d_i, l_i]$ that construct $\Sigma_i$ through Eq. \eqref{eq:ldl}. Together, they form the Gaussian distribution \(\mathcal{N}(\boldsymbol{\mu}_i, \mathbf{\Sigma}_i)\), providing both an accurate state estimate and a calibrated measure of its uncertainty. 


\subsection{Objective Functions}

Although mean and covariance can be jointly optimized as in \cite{de2020simultaneously}, we empirically found that decoupled optimizations lead to more stable training and better convergence.

\textit{Mean (bias) loss}:
Following DPC‑Net~\cite{peretroukhin2017dpc}, we minimize the geodesic distance between the bias‑corrected estimate and ground-truth:
\begin{equation}
\begin{aligned}
L_{\mathrm{geo}} &= \frac{1}{2T}\sum_{i=0}^{T}\!\bigl\|\log\bigl(\exp(\hat{\boldsymbol\xi}_i)\,\mathbf T_{\mathrm{gt},i}^{-1}\bigr)\bigr\|^{2}_{\mathbf P}
              - \bigl\|\log(\mathbf T_{\mathrm{gt},i}^{-1})\bigr\|^{2}_{\mathbf P} \\ &+\lambda_s\mathcal L_s,
\end{aligned}\end{equation} where $\hat{\boldsymbol\xi}_i$ is the network prediction of error, and $\|\mathbf v\|^{2}_{\mathbf P}=\mathbf v^{\top}\mathbf P\,\mathbf v$ with an empirical weighting matrix $\mathbf P$. For training Mamba, we introduce an additional smoothness penalty, denoted as $\mathcal{L}_s=\dfrac{1}{T-1}\sum_{i=1}^T(\Delta\log(\mathbf T_{\mathrm{gt},i}^{-1}) - \Delta\log\bigl(\exp(\hat{\boldsymbol\xi}_i)\,\mathbf T_{\mathrm{gt},i}^{-1}))$, where $\Delta(\cdot)$ denotes the change in pose between the previous and current states, to capture the smoothness of the trajectory.

\textit{Covariance loss}:
To calibrate uncertainty, we minimize the negative log‑likelihood (NLL):
\begin{equation}
\begin{aligned}
L_{\mathrm{nll}}= & \sum_{i=0}^{T} -\log \text{I\kern-0.15em P}(\boldsymbol\xi_i \mid (\boldsymbol\mu, \boldsymbol\Sigma)_i)
\label{eq:nll}
\end{aligned}
\end{equation}
where $\text{I\kern-0.15em P}$ is probability measure, $\boldsymbol\xi_i$ is the observed error, and $(\boldsymbol\mu, \boldsymbol\Sigma)_i$ is the network prediction of error distribution. The total objective is $\mathcal L=L_{\mathrm{geo}}+L_{\mathrm{nll}}$ and is optimised end‑to‑end with AdamW.

\subsection{Implementation Details} We fix the sequence length $T=100$ and truncate the input data accordingly. For the dimensions of the input features, we set $d_\text{image}=256$, $ d_\text{imu}=128$, and $ d_\text{odom}=128 $. Additionally, we employ a skip connection from the raw VIO output to the decoder, which is added element-wise to the output mean through a shallow MLP of dimension 6. Both the final decoder and the skip connection MLP are zero-initialized for training stability. For training parameters, we set $\lambda_s=100$. We train all our models on 4xNVIDIA RTX 6000 GPUs with an effective batch size of 128 (32 per GPU) and learning rate set to $1e-6$ for mean learning and $1e-4$ for covariance learning. The training time of our model is around 1 hour.


\section{EXPERIMENTS \& RESULTS}
\label{sec:exp}

We evaluate MUSE against other SOTA uncertainty estimation methods for pose estimation on various odometry methods. Moreover, we demonstrate our method's ability to estimate well-calibrated uncertainty in both zero-mean and non-zero-mean settings when compared with different baselines. Finally, we showcase the robustness of MUSE on the challenging environments in UnCal-Flight dataset.

\subsection{Datasets}
\subsubsection{EuRoC Dataset} The EuRoC dataset \cite{burri2016euroc} is a widely used benchmark for evaluating VO/VIO and SLAM algorithms. It consists of multiple sequences recorded in indoor environments using a Micro Aerial Vehicle (MAV) equipped with a synchronized stereo camera and an inertial measurement unit (IMU). The dataset provides precise ground-truth trajectories obtained through a motion capture system, making it suitable for assessing pose estimation accuracy and uncertainty quantification methods.

We use the Machine Hall trajectories from the EuRoC dataset, with \texttt{MH\_01} to \texttt{MH\_04} for training and \texttt{MH\_05} for evaluation. We run MSCKF \cite{sun2018robust}, VINS-Fusion \cite{qin2019a}, ORB-SLAM 3 \cite{campos2021orb}, and DPVO \cite{teed2023deep} to obtain their respective estimations.

\begin{figure} [t]
    \centering
    \includegraphics[width = \columnwidth]{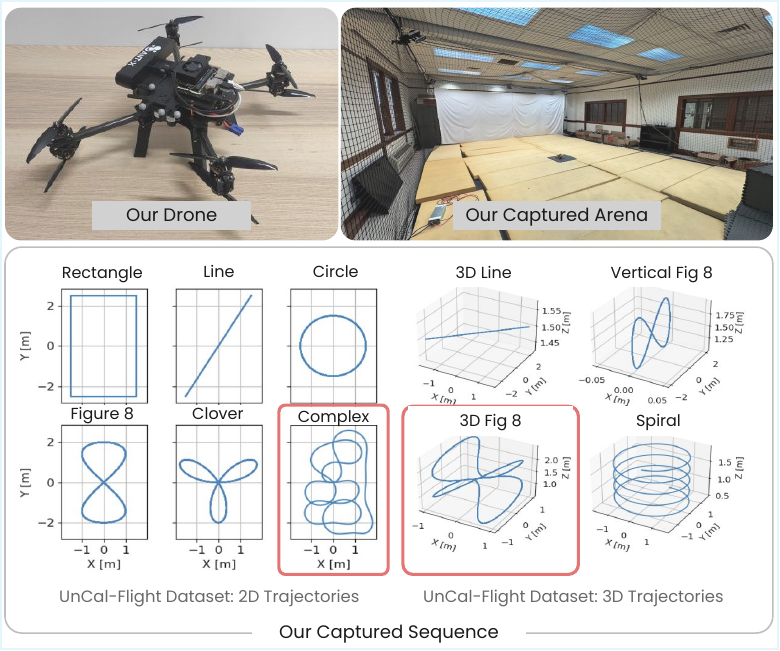}
    \caption{Our drone, captured arena, and trajectories included in UnCal-Flight dataset. Red denotes evaluation trajectories.}
    \label{fig:shapes of trajectories}
    \vspace{-0.5cm}
\end{figure}

\subsubsection{UnCal-Flight Dataset} While the EuRoC dataset is valuable, its number and diversity of trajectories are limited. To address this limitation, we collected an uncalibrated flight (UnCal-Flight) dataset, featuring more complex motion patterns in varied environments. We used ROG-X \cite{rogx} to gather UAV-based visual-inertial data. The drone is equipped with a ZED 2i stereo camera that captures RGB images at 15 Hz and an onboard IMU recording at 200 Hz. The flight arena, measuring $3\times 5 \times 7 \, m^3$, is equipped with a Vicon motion capture system providing accurate ground-truth measurements at 100 Hz.

To increase dataset diversity, we conducted drone flights along various trajectories at different speeds and with dynamic yaw motions inspired by \cite{antonini2018blackbird}, under varying lighting conditions and with a human moving in and out of the sensor’s frame. The shapes of the
collected trajectories are illustrated in Fig.~\ref{fig:shapes of trajectories}, with minor deviations due to safety margins and tracking performance. In total, 145 trajectories were collected.

\begin{table*}[t]
\centering
\resizebox{\textwidth}{!}{
    \begin{tabular}{lcccccccc}
        \toprule
        \multirow{2}{*}{\textbf{Methods}} 
        & \multicolumn{2}{c}{\textbf{VINS-Fusion}} 
        & \multicolumn{2}{c}{\textbf{MSCKF}} 
        & \multicolumn{2}{c}{\textbf{ORB-SLAM 3}} 
        & \multicolumn{2}{c}{\textbf{DPVO}} \\
        \cmidrule(lr){2-3} \cmidrule(lr){4-5} \cmidrule(lr){6-7}  \cmidrule(lr){8-9} 
        & RMSE [m]$\downarrow$ & GEO [rad]$\downarrow$
        & RMSE [m]$\downarrow$ & GEO [rad]$\downarrow$
        & RMSE [m]$\downarrow$ & GEO [rad]$\downarrow$
        & RMSE [m]$\downarrow$ & GEO [rad]$\downarrow$\\
        \midrule
        Raw Odom & 0.164 & 0.014 & 0.063 & \bf  0.008 & \bf  0.027 & \bf  0.002 & \bf  0.048 & \bf 0.005\\
        D-DICE \cite{de2020simultaneously} & 0.161 & 0.020 & 0.066 & 0.010 & 0.032 & 0.010 & 0.070 & 0.020  \\
        DPC-Net \cite{peretroukhin2017dpc} & 0.162 &  0.012 & 0.063 & \bf 0.008 & \bf 0.027 & \bf 0.002 & \bf  0.048 & \bf  0.005 \\
        MUSE (Ours) & \bf 0.089 & \bf 0.009 & \bf 0.062 & \bf 0.008 &  0.028 & \bf 0.002 & \bf 0.048 & \bf  0.005 \\
        \bottomrule
    \end{tabular}
}
\caption{Pose correction on EuRoC.}
\label{tab:pose_corr_euroc}
\vspace{-2mm}
\end{table*}

\begin{table*}[t]
\centering
\resizebox{0.75\textwidth}{!}{
    \begin{tabular}{lcccccccc}
        \toprule
        \multirow{2}{*}{\textbf{Methods}} & \multicolumn{2}{c}{\textbf{VINS-Fusion}} & \multicolumn{2}{c}{\textbf{MSCKF}} & \multicolumn{2}{c}{\textbf{ORB-SLAM 3}} & \multicolumn{2}{c}{\textbf{DPVO}} \\
        \cmidrule(lr){2-3} \cmidrule(lr){4-5} \cmidrule(lr){6-7}  \cmidrule(lr){8-9}  
        & LL$\uparrow$ & ENCE$\downarrow$ & LL$\uparrow$ & ENCE$\downarrow$ & LL$\uparrow$ & ENCE$\downarrow$ & LL$\uparrow$ & ENCE$\downarrow$ \\
        \midrule
        \multicolumn{9}{c}{\textit{{Zero-Mean Uncertainty Quantification}}} \\
        Empirical & \underline{36.26} & \underline{0.497} & 25.08 & 0.904 & \bf 57.03 & \bf 0.080 & \underline{48.92} & \underline{0.135}\\
        DICE \cite{liu2018deep} & 26.11 & 0.765 & \underline{33.54} & \underline{0.439} & \underline{44.97} & 0.432 & 37.00 & 0.473\\
        D-DICE \cite{de2020simultaneously} & 21.74 & 0.704 & 28.34 & 0.611 & 38.08 & 0.561 & 31.14 & 0.601 \\
        MUSE (Zero-Mean) & \bf 37.95 & \bf 0.300 & \bf 44.33 & \bf 0.138 & \bf 57.03 & \underline{0.097} & \bf{49.09} & \bf{0.123} \\
        \midrule
        \multicolumn{9}{c}{\textit{{Non-Zero-Mean Uncertainty Quantification}}} \\
        D-DICE \cite{de2020simultaneously} & 21.33 & 0.331 & 28.69 & 0.500 & 31.46 & 0.639 & 24.06 & 0.646 \\
        MUSE (Ours) & \bf 40.46 & \bf 0.278 & \bf 43.51 & \bf 0.145 & \bf 54.83 & \bf 0.087 & \bf 49.06 & \bf 0.239\\
        \bottomrule
    \end{tabular}
}
\caption{Uncertainty quantification on EuRoC.}
\label{tab:uq}
\vspace{-2mm}
\end{table*}

\begin{table*}[t]
\centering
\resizebox{0.95\textwidth}{!}{
    \begin{tabular}{lcccccccc}
        \toprule
        \multirow{2}{*}{\textbf{Methods}} 
        & \multicolumn{4}{c}{\textbf{Yaw Constant}} 
        & \multicolumn{4}{c}{\textbf{Yaw Forward}} \\
        \cmidrule(lr){2-5} \cmidrule(lr){6-9}
        & RMSE [m]$\downarrow$ & GEO [rad]$\downarrow$
        & LL$\uparrow$ & ENCE$\downarrow$
        & RMSE [m]$\downarrow$ & GEO [rad]$\downarrow$
        & LL$\uparrow$ & ENCE$\downarrow$\\
        \midrule
        Raw Odom & 0.045 & \bf 0.022 & - & - & 0.065 & 0.042 & - & -\\ 
        D-DICE \cite{de2020simultaneously} & 0.044 & \bf 0.022 & 24.36 & 0.631 & 0.062 & 0.042 & 22.13 & 0.522 \\
        DPC-Net \cite{peretroukhin2017dpc} & 0.045 & \bf 0.022 & - & - & 0.059 & 0.041 & - & - \\
        MUSE (Ours) & \bf 0.035 & \bf 0.022 & \bf 32.00 & \bf 0.421 & \bf 0.040 & \bf 0.033 & \bf 29.46 & \bf 0.379 \\
        \bottomrule
    \end{tabular}
}
\caption{Pose correction and uncertainty quantification on UnCal-Flight Dataset. Raw Odom is obtained from ZED-VO. }
\label{tab:lab_eval}
\vspace{-4mm}
\end{table*}

\subsection{Evaluation Setup}
\subsubsection{Evaluation Settings}
We evaluate our methods on two settings: (1) zero-mean and (2) non-zero-mean uncertainty estimation. In the first setting, we assume the mean pose error to be zero and modify our decoder to output only the covariance. This setting is valuable for well-tuned and highly accurate pose estimation methods such as ORB-SLAM 3 \cite{campos2021orb} or DPVO \cite{teed2023deep}. The second setting is to evaluate the performance of our pose correction, which is valuable for under-performing odometry methods or challenging scenarios. To better evaluate the heteroscedasticity of the covariance estimation, instead of computing the metrics on a frame-to-frame basis, we split the poses into 100-frame overlapping chunks and evaluate the metrics on a chunk-to-chunk basis. 

\subsubsection{Baselines} We compare our method against state-of-the-art uncertainty quantification approaches, including DICE \cite{liu2018deep}, D-DICE \cite{de2020simultaneously}, and DPC-Net \cite{peretroukhin2017dpc}. We implement DICE and D-DICE with parameters that best follow their respective papers. The images are resized to $48 \times 64$ for DICE training, and to $80 \times 128$ for D-DICE and DPC-Net. In the zero-mean setting, we include the empirical uncertainty of each VO/VIO model. For the MSCKF, we utilize the covariance obtained from the filter, while for others, we fit a Gaussian distribution at each timestep throughout the sample of chunks. Though simple, we observe that this baseline is quite competitive in capturing the heteroscedastic nature of the covariance. Additionally, to evaluate the impact of multimodal input on uncertainty estimation, we train our network using different input configurations: (i) image-only, (ii) image with IMU, (iii) image with estimations from VO/VIO models, and (iv) full multimodal input incorporating all of the above. This allows us to analyze the contribution of each modality to the overall uncertainty quantification. 
\subsection{Metrics} 
\paragraph{Uncertainty Calibration} We evaluate calibration using Log-Likelihood (LL) and Expected Normalized Calibration Error (ENCE) \cite{levi2022evaluating}. Similar to Expected Calibration Error (ECE) in classification, ENCE measures the mismatch between observed errors and predicted uncertainty in regression tasks. Low ENCE indicates well-calibrated confidence, while high ENCE suggests over- or under-confidence.

For uncertainty quantification, we define the uncertainty score over 6D pose vector as $u_t = \sqrt{\mathrm{tr}(\boldsymbol{\Sigma}_t)}$. Samples are grouped into equal-sized M bins $\{B_j\}_{j=1}^M$, and we compute the root mean variance $\mathrm{RMV}(j) = \sqrt{\frac{1}{|B_j|}\sum_{t\in B_j} u_t^{\,2}}$ and the empirical root mean square error (RMSE) as $\mathrm{RMSE}(j) = \sqrt{\frac{1}{|B_j|}\sum_{t\in B_j}\|\boldsymbol{\xi}_t-\boldsymbol{\mu}_t\|^2}$.
Then, ENCE is computed as:
$
    \mathrm{ENCE} = \dfrac{1}{M}\sum_{j=1}^M \dfrac{|\mathrm{RMSE}(j)-\mathrm{RMV}(j)|}{\mathrm{RMV}(j)}.
$

\paragraph{Pose Correction} For the mean values, which correspond to the corrected pose estimation, we use the Root Mean Square Error (RMSE) for the translational error and Geodesic Distance (GEO) for the rotational error $\text{GEO($R_{est}, R_{gt})$}=\arccos\left(\frac{\mathrm{tr}(R_{est} R_{gt}^T) - 1}{2}\right).$



\begin{table}[t]
\centering
\resizebox{\columnwidth}{!}{
    \begin{tabular}{cccccccc}
        \toprule
        \multicolumn{3}{c}{\textbf{Input}} 
        & \multicolumn{4}{c}{\textbf{Yaw Forward}} \\
        \cmidrule(lr){1-3} \cmidrule(lr){4-7}
        Cam & IMU & Odom 
        & RMSE [m]$\downarrow$ & GEO [rad]$\downarrow$
        & LL $\uparrow$ & ENCE $\downarrow$ \\
        \midrule
        \checkmark & & & 0.061 & 0.043 & 26.13 & 0.677\\
        \checkmark & \checkmark & & 0.058 & 0.043 & 25.62 & 0.628\\
        \checkmark & & \checkmark & 0.050 & 0.038 & 26.25 & \bf 0.359\\
        \checkmark & \checkmark & \checkmark & \bf 0.040 & \bf 0.033 & \bf 29.46 & 0.379\\
        \bottomrule
    \end{tabular}
}
\caption{Ablation of multimodal inputs.}
\label{tab:ablation}
\vspace{-7mm}
\end{table}

\begin{figure*}
    \centering
    \includegraphics[width=\textwidth]{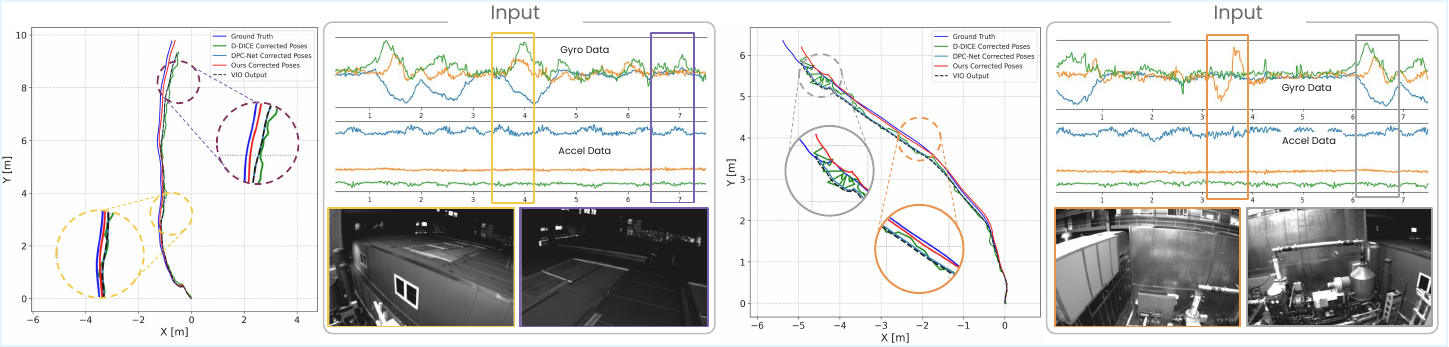}
    \caption{Qualitative results of pose correction performance on EuRoC Dataset (\texttt{MH\_05\_difficult}). MUSE (red), which utilizes visual, inertial, and odometry signals as inputs, is able to accurately detect and correct pose drifts (zoomed in regions), whereas DPC-Net (light blue) and D-DICE (green) can only support visual input and thus fail.}
    \label{fig:euroc_pose_corr}
    \vspace{-5mm}
\end{figure*}

\subsection{Pose Corrections}
In Table \ref{tab:pose_corr_euroc}, we present our results on pose correction compared with baseline methods on the same set of odometry. For odometry methods with relatively higher raw odometry errors, such as VINS-Fusion, our method is able to decrease its RMSE by 45.7\%, whereas other baselines show minor improvement or regression. For other methods (e.g., MSCKF, ORB-SLAM, and DPVO), their raw odometry has already obtained relatively low pose errors thanks to their well-tuned parameters on EuRoC. Our pose correction shows no regression in performance, whereas D-DICE degrades the raw odometry's performance. 

Moreover, we present qualitative results in Fig. \ref{fig:euroc_pose_corr}. Here, we zoom in on the portion of the trajectory where VIO starts to drift from the GT, and provide the according sensor input. Thanks to our utilization of multimodal inputs, our method is able to detect and correct VIO poses, whereas single-model methods such as DPC-Net and D-DICE fail to. This further validates our intuition outlined in Fig. \ref{fig:multimodal} about the importance of multimodality in capturing pose drifts and uncertainty.  

\subsection{Uncertainty Quantification}

We present our uncertainty quantification performance for both zero-mean and non-zero-mean settings in Table \ref{tab:uq}. 


\subsubsection{Zero-Mean Uncertainty Quantification} Under zero-mean settings, we compare MUSE with DICE \cite{liu2018deep}, D-DICE \cite{de2020simultaneously}, and empirical covariance.  Our method outperforms all the baselines in estimating the uncertainty of VINS-Fusion, MSCKF, and DPVO poses, and achieves competitive results on ORB-SLAM 3 with the empirical covariance. Notably, DICE and D-DICE fall short of empirical uncertainty in this setting, which may follow from the lack of sequential and multimodal modeling in their methods. 

\subsubsection{Non-zero-mean Uncertainty Quantification} Under non-zero-mean settings, since DPC-Net does not predict uncertainty, we only compare MUSE with D-DICE. Our uncertainty metrics are better across the board from D-DICE.

\subsection{Challenging Scenarios} 
\begin{figure*}
    \centering
    \includegraphics[width=\textwidth]{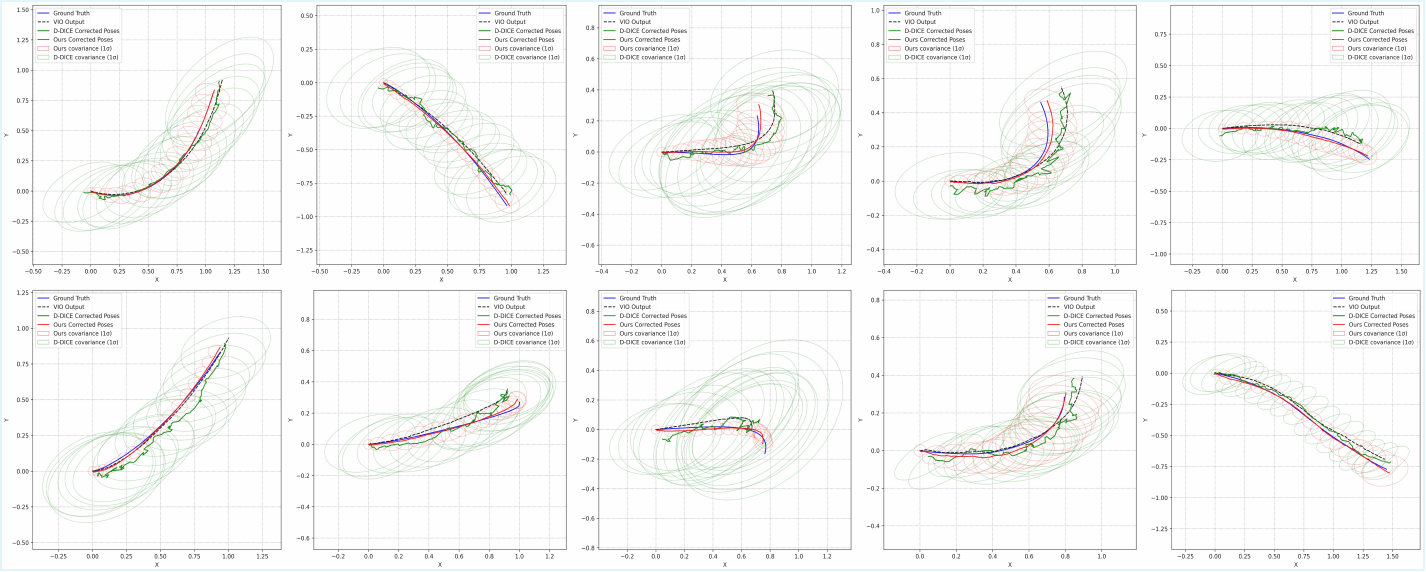}
    \caption{Qualitative Results on UnCal-Flight Dataset. Our method can predict accurate pose correction (red curve) with reasonable uncertainty (red circles) compared to D-DICE (green curve and green circles). Moreover, our pose correction shows significant improvement over raw VIO odometry (black dotted lines).}
    \label{fig:lab_qual}
    \vspace{-5mm}
\end{figure*}

Under this setting, we want to evaluate our performance on the challenging UnCal-Flight Dataset. For the odometry, we use the proprietary odometry algorithm provided by the onboard ZED 2i camera, which we call ZED-VO. We report both pose corrections and non-zero-mean uncertainty quantification in Table \ref{tab:lab_eval}. Our method outperforms all baselines in pose correction for both yaw-constant and yaw-forward subsets. Notably, in terms of RMSE metrics, our method achieves 22.2\% and 38.4\% improvement over the raw odometry on the above two subsets, respectively. For uncertainty quantification, our method outperforms D-DICE in both LL and ENCE metrics in both subsets. Moreover, we showcase the qualitative results in this setting in Fig. \ref{fig:lab_qual}, with uncertainty plotted. Our method provides better correction with reasonable uncertainty compared to D-DICE. The average inference time for a single window with all modalities as input is under 3 ms on a single NVIDIA RTX 6000 GPU.

\subsection{Ablative Studies}

To better understand the importance of multimodal input when learning the error distribution, we conduct an ablative study comparing our design with different combinations of input modalities and report the numbers in Table \ref{tab:ablation}. We observe that using all modalities (camera, IMU, and estimated odometry) usually leads to the best performance in both pose correction and uncertainty estimation. 

\section{CONCLUSIONS}
\label{sec:conclusion}
In this paper, we introduce MUSE, a novel framework for multimodal uncertainty quantification for any black-box pose estimation method. Using a highly efficient and powerful state-space model, MUSE is able to estimate the error distribution of a given odometry with well-calibrated uncertainty, where the predicted mean can be used to accurately correct the pose estimation. Experiments on both EuRoC and the challenging UnCal-Flight dataset demonstrate the effectiveness of our approach across diverse odometry methods and sensing conditions. Beyond improved calibration and correction, our ablation results highlight the importance of multimodal integration, showing consistent gains when fusing visual, inertial, and odometry streams. In future work, we plan to extend MUSE to additional sensing modalities and integrate into downstream planning and control frameworks, moving toward a fully uncertainty-aware autonomy stack.


\section*{ACKNOWLEDGMENT}
This work is supported by the Air Force Office of Scientific Research Grant (AFOSR) Grant AF FA9550-25-1-0274, the National Aeronautics and Space Administration (NASA) under Grant 80NSSC22M0070, and by the National Science Foundation (NSF) under Grants CMMI-2135925, CPS-2311085, IIS-2331878, IIS-2331879, IIS-2340254, IIS-2312102, CNS-2414227, IIS-2404385, and CCF-2525287. Henry Che is supported by the NSF GRFP fellowship.




\bibliographystyle{IEEEtran}
\bibliography{ref}

\end{document}